\pdfoutput=1

\documentclass[11pt]{article}

\usepackage[preprint]{acl}

\usepackage{times}
\usepackage{latexsym}

\usepackage[T1]{fontenc}

\usepackage[utf8]{inputenc}

\usepackage{microtype}

\usepackage{inconsolata}

\usepackage{graphicx}

\usepackage{listings}
\usepackage{booktabs}
\usepackage{multirow}
\usepackage{xcolor}

\lstdefinestyle{mystyle}{
    backgroundcolor=\color{white},
    keywordstyle=\color{magenta},
    numberstyle=\tiny\color{gray},
    stringstyle=\color{purple},
    basicstyle=\ttfamily\footnotesize,
    breakatwhitespace=false,         
    captionpos=b,                    
    keepspaces=true,                 
    numbers=left,                    
    numbersep=5pt,                  
    showspaces=false,                
    showstringspaces=false,
    showtabs=false,
}
\title{Large Language Models Are Overparameterized Text Encoders }

\author{Thennal D K \\ IIIT Kottayam \\  \texttt{thennal21bcs14@iiitkottayam.ac.in}
     \And  
     Tim Fischer \\ University of Hamburg \\ \texttt{tim.fischer@uni-hamburg.de} 
     \AND
     Chris Biemann \\ University of Hamburg \\ 
     \texttt{biemann@informatik.uni-hamburg.de}
     }

\begin{document}
\maketitle
\begin{abstract}
Large language models (LLMs) demonstrate strong performance as text embedding models when finetuned with supervised contrastive training. However, their large size balloons inference time and memory requirements. In this paper, we show that by pruning the last $p\%$ layers of an LLM before supervised training for only 1000 steps, we can achieve a proportional reduction in memory and inference time. We evaluate four different state-of-the-art LLMs on text embedding tasks and find that our method can prune up to 30\% of layers with negligible impact on performance and up to 80\% with only a modest drop. With only three lines of code, our method is easily implemented in any pipeline for transforming LLMs to text encoders. We also propose L\textsuperscript{3}Prune, a novel layer-pruning strategy based on the model's initial loss that provides two optimal pruning configurations: a large variant with negligible performance loss and a small variant for resource-constrained settings. On average, the large variant prunes 21\% of the parameters with a $-0.3$ performance drop, and the small variant only suffers from a $-5.1$ decrease while pruning 74\% of the model. We consider these results strong evidence that LLMs are overparameterized for text embedding tasks, and can be easily pruned.

\end{abstract}

\section{Introduction}
\label{sec:introduction}

In the past few years, the field of natural language processing (NLP) has seen a significant shift towards large-scale language models (LLMs). These models, due to a combination of their large size, extensive pre-training, and instruction-following ability, have achieved state-of-the-art performance on a wide range of NLP tasks, such as language modeling, text generation, and text understanding \citep{dubey2024llama3herdmodels,brown-etal-gpt3, jiang2023mistral7b}. 

Despite their strong generative capabilities, decoder-only LLMs have seen comparatively little adoption for text embedding tasks until recently \citep{behnamghader_llm2vec_2024}. Text embedding, which involves mapping a text sequence of varying length to a fixed-dimensional vector representation, is a fundamental task in NLP and is used as a building block for a wide range of downstream tasks, such as semantic textual similarity, information retrieval, and text classification. Further, it is a fundamental step required for retrieval-augmented generation, a well-known paradigm for improving the performance of LLMs in knowledge-intensive tasks \citep{lewis2020retrieval}. Traditionally, text embedding models have been based on masked language models (MLMs) and bidirectional encoders, such as BERT \citep{devlin-2019-BERT} and T5 \citep{raffel-etal-2020-t5}, typically adapted for text embedding tasks by following a multi-step training pipeline consisting of weakly- and fully-supervised contrastive training \citep{ni-etal-2022-GTR,li-etal-2023-gte,shitao-etal-2023-bge}.

Decoder-only LLMs, however, offer several advantages over their encoder-only counterparts. They are more sample-efficient during pre-training, leverage instruction-following capabilities for task generalization, and benefit from a rich and evolving research ecosystem \citep{clark2020electra, asai-etal-2023-Tart, behnamghader_llm2vec_2024}. Further, the availability of high-performing public pre-trained LLMs and their continual development make it appealing to explore their use for text embedding tasks. To this end, several studies have experimented with various pipelines, training methods, and architectural modifications, effectively converting LLMs into state-of-the-art text embedding models with small amounts of supervised contrastive training \citep{behnamghader_llm2vec_2024, li-li-2024-bellm, Ma2023-repllama, Muennighoff2022-SGPT, springer_repetition_2024, lee2024nvembedimprovedtechniquestraining}.

On the other hand, the increasingly large size of LLMs, with parameters ranging up to 540B \citep{brown-etal-gpt3,chowdhery2022palm,dubey2024llama3herdmodels}, stands in stark contrast to traditional small bidirectional encoders of sizes almost universally less than 1B parameters \citep{li-etal-2023-gte,shitao-etal-2023-bge}. Even the smallest LLMs in use typically have 3-8B parameters \cite{abdin2024phi3technicalreporthighly}. Consequently, inference with LLM-based text encoders is far more demanding in terms of compute and memory requirements in comparison to traditional methods.   

Therefore, there are a variety of post-training techniques for reducing the cost of LLMs, such as pruning, quantization, and distillation \citep{zhu2023surveymodelcompression}. In particular, the recent work of \citet{gromov_unreasonable_2024} has shown that LLMs can be pruned to up to half their size with minimal impact on downstream performance (i.e. question answering) by dropping the last half of the model's layers, with the exception of the final layer, and applying a small amount of parameter-efficient finetuning. Layer-dropping as a pruning strategy has particular benefits: it is straightforward to implement, with memory and inference time decreasing linearly with the number of layers dropped, and it can be combined with other efficiency methods such as quantization.


In this work, we build on these findings and apply them in the context of text embedding, resulting in a easy-to-use and efficient approach to transform any pre-trained decoder-only LLM into a much smaller text embedding model. By simply pruning the last $n\%$ layers of a model before supervised contrastive training, we reduce the final model size with a proportional decrease in memory and inference time. We experiment with four different decoder-only LLMs ranging from 3.8B to 8B parameters with a variety of pruning percentages and show that up to 30\% of a model's layers may be pruned with almost no impact in performance, and may even \emph{increase} it. Even intensive pruning on the order of 80\% still provide reasonably effective text embedding models, with a drop in performance on the downstream task from $64.9$ to $59.8$ for our highest-performing model.

Further, we propose \textbf{L\textsuperscript{3}Prune}, a simple and novel method that pinpoints particular layers to prune to based on the initial without requiring significant testing or experimentation. With no input, our method produces both \emph{a)} a lightly-pruned model, 69-89\% of the original size with minimal performance loss of $-0.2$ on average and even a performance \emph{improvement} in one model, and \emph{b)} a heavily pruned model, 16-36\% of their original size with a modest performance drop of $-4.4$ to $-6.9$. 

We note that our pruning technique is both orthogonal to other commonly used pruning and quantization methods and to the specifics of the methodology used to finetune the LLM for text embedding. As such, it can be seamlessly integrated with other efficiency techniques and finetuning methods to further optimize model size and performance. 

Our contributions can be summarized as follows:
\begin{itemize}
    \item We are the first to apply pruning in a text embedding setting, formulating a simple procedure that can be easily applied to pipelines converting an LLM to a text encoder.
    \item We demonstrate that LLMs can be pruned by up to 30\% with negligible impact on the quality of representations and up to 80\% with a modest performance drop. 
    \item We propose and evaluate L\textsuperscript{3}Prune, a novel method that identifies layers to prune by leveraging the model's initial loss, thus minimizing the need for trial-and-error for effective pruning.
\end{itemize}

Overall, our work demonstrates that decoder-only LLMs are generally overparameterized for text embedding tasks, and that significant reductions in model size can be achieved with minimal impact on performance. We release the full code for L\textsuperscript{3}Prune \footnote{\url{https://github.com/thennal10/l3prune}}. 

\section{Related Work}
\label{sec:related-work}

\subsection{Encoder-only Text Embedding Models}

BERT-based models have largely dominated the field of text representation in the past, relying on supervised training with natural language inference or sentence similarity to produce high-quality sentence embeddings \citep{conneau-etal-2017-supervised,reimers-gurevych-2019-sbert}. Recent methods have further improved these representations through large-scale contrastive pretraining followed by multi-task fine-tuning \citep{ni-etal-2022-GTR,wang-etal-2022-e5,li-etal-2023-gte,shitao-etal-2023-bge}. These methods generally require a complex multi-stage training pipeline that demands substantial engineering effort, along with large-scale compute-intensive pretraining \citep{zhang-etal-2024-e5}.

\subsection{Decoder-only Text Embedding Models} 

A variety of recent works have explored leveraging LLMs and their capabilities to generate high-quality text representations. Generally, a combination of \emph{(a)} a pooling method, \emph{(b)} architectural modifications, and \emph{(c)} supervised or unsupervised fine-tuning is used to effectively convert LLMs to text embedding models.

The majority of prior work consider two straightforward pooling strategies to extract embeddings for a sequence of tokens: mean pooling and last-token pooling \cite{springer_repetition_2024, jiang-etal-2023-prompteol, behnamghader_llm2vec_2024, Muennighoff2022-SGPT, wang-etal-2024-improving-text}. Mean pooling is more effective with bidirectional embedding models \citep{behnamghader_llm2vec_2024, wang-etal-2022-e5} while last-token pooling is generally preferred when working with causal attention \citep{lee2024nvembedimprovedtechniquestraining, behnamghader_llm2vec_2024}. \citet{Muennighoff2022-SGPT} introduces weighted mean pooling, assigning a higher weight to later tokens to offset the autoregressive nature of decoder-only LLMs, with significant success. \citet{lee2024nvembedimprovedtechniquestraining} utilizes a trainable latent attention layer as a pooling technique and obtains consistent improvement. 

Several studies identify the causal attention mechanism of decoder-only LLMs as a obstacle in obtaining performant representations, and suggest modifications to the architecture to compensate. \citet{li-li-2024-bellm} and \citet{behnamghader_llm2vec_2024} replace the causal attention mechanism with bidirectional attention. \citet{muennighoff2024generative} utilizes a hybrid objective with both bidirectional representation learning and causal generative training. \citet{lee2024nvembedimprovedtechniquestraining} finds that simply removing the causal attention mask works compellingly well.

Finally, both supervised and unsupervised finetuning have been extensively explored to significantly improve the performance of decoder-only LLMs in representational tasks, with supervised training consistently producing the best results \citep{behnamghader_llm2vec_2024, Muennighoff2022-SGPT, jiang-etal-2023-prompteol}. Several modifications to the training pipeline have been proposed, such as an additional masked token prediction training step \citep{behnamghader_llm2vec_2024}, or a two-stage instruction-tuning setup \citep{lee2024nvembedimprovedtechniquestraining}. The zero-shot setting has also been studied with limited success by \citet{springer_repetition_2024} and \citet{jiang-etal-2023-prompteol}. 


\subsection{LLM Pruning}

\begin{figure*}[t]
  \includegraphics[width=\textwidth]{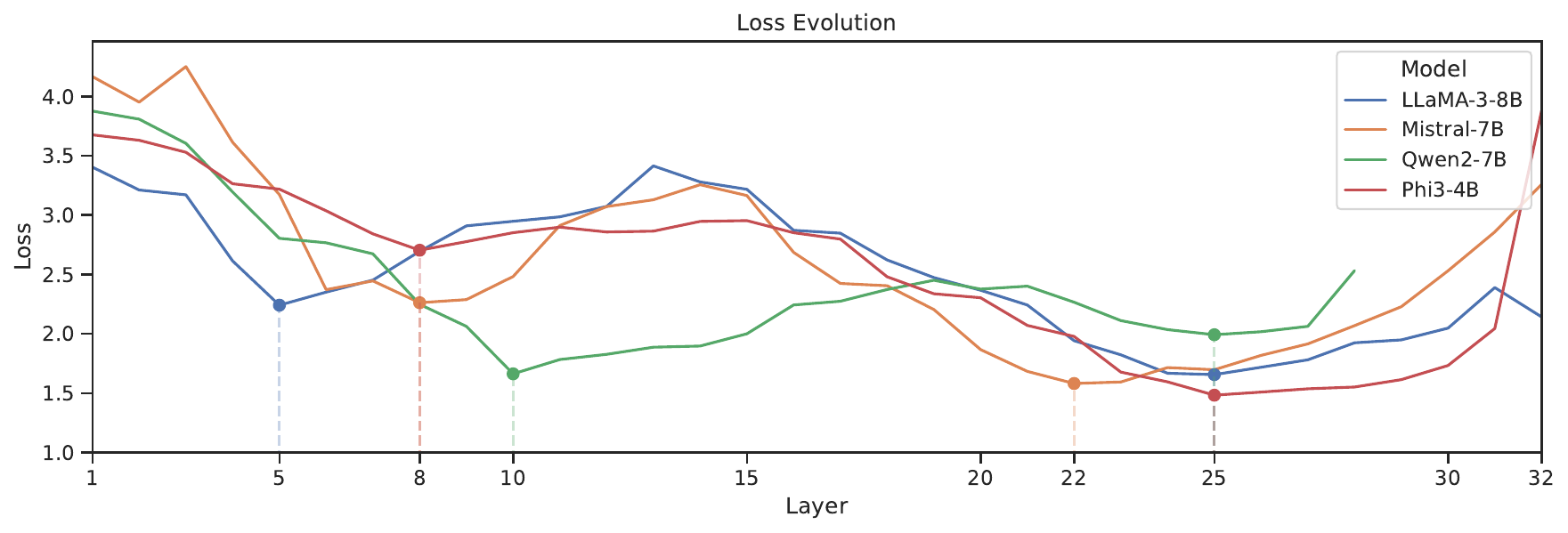}
  \caption{
  The loss values extracted from the layerwise embeddings of samples from the training dataset. These values were obtained for each unmodified model. The marked points indicate the layer with minimal loss before and after the midpoint.}
  \label{fig:loss_evo}
\end{figure*}

Pruning as a method of size reduction has a long history in the field of deep learning \citep{cheng2024surveypruning}. Classic pruning techniques sparsify networks by removing individual parameters based on various criteria \citep{lecun1990optimal, han2015learning}. While these models were smaller, these techniques generally lead to irregular sparsification patterns that require specialized hardware or libraries to fully utilize. Structured pruning techniques were developed to remove irrelevant groups of parameters together, such as particular channels or filters in convolutional neural networks \citep{wen2016learningstructured, li2022pruning}.

Recent work has focused on applying structure pruning methods to transformers. Almost every possible component of the model architecture is studied as candidates for removal, most prominently methods that drop attention heads \citep{voita2019analyzing, michel2019sixteen, kim2020fastformers} and layers \cite{fan2021layer, zhang-etal-2022-pcee, sajjad2023effect, gromov_unreasonable_2024, men2024shortgptlayerslargelanguage, fan_not_2024}. Prior literature on layer pruning generally consider BERT-like models \citep{fan2021layer, sajjad2023effect}, with recent studies shifting focus to decoder-only LLMs \citep{gromov_unreasonable_2024, men2024shortgptlayerslargelanguage, fan_not_2024}. 

\citet{sajjad2023effect} finds that for BERT-like models, dropping the last layers is the best layer pruning strategy. \citet{gromov_unreasonable_2024} extends this research to decoder-only LLMs, and presents a layer pruning strategy, pruning a block of layers based on angular distance between layer representations. Their results indicate that the last layer in particular is essential for maintaining performance. Informed by this finding, they propose a simpler strategy: dropping the last $n$ layers with the exception of the final layer. They conclude that simply dropping the last layers works effectively to prune the model, with a caveat: after dropping the layers, it is required to "heal" the model via finetuning with QLoRA \citep{dettmers2024qlora} for 1000 steps. 

While these results suggest that the last layer in particular is essential when pruning LLMs for text generation, this is not necessarily the case when utilizing the LLM for other tasks. To this end, \citet{fan_not_2024} finds that for "simpler" tasks such as sentiment analysis, early stopping—stopping the inference after a certain number of layers—is an effective strategy to significantly reduce inference time with minimal impact on performance. The authors suggest that the later layers of LLMs, including the final layer, may not be necessary when using the LLM representations for other tasks. 

\section{Pruning}

We borrow the intuition from \citet{gromov_unreasonable_2024}, that the representations in a transformer can be thought of as a slowly changing function of the layer index. Specifically, the representation can be formulated as the following iterative residual equation:

\begin{equation}
    x^{(\ell+1)} = x^{(\ell)} + f(x^{(\ell)}, \theta^{(\ell)}),
\end{equation}
where \(x^{(\ell)}, \theta^{(\ell)}\), respectively, are the multi-dimensional input and parameter vectors for layer \(\ell\), and \(f(x, \theta)\) describes the transformation of one multi-head self-attention and MLP layer block.

The authors assert that these representations converge to a slowly changing function:
\begin{equation}
    x^{(\ell)} \approx x^{(\ell-1)} + \epsilon
\end{equation}

with $\epsilon \ll x^{\ell}$ as an approximation. They verify this hypothesis experimentally by calculating the distance between layer representations and using them for a pruning algorithm. Their findings indicate that the earlier layers have a significantly larger impact on the representation compared to the later layers, with a particular caveat: the final layer also modifies the representation significantly. They thus propose and verify a simpler pruning strategy, where the last $n$ layers of the model, excluding the final layer, are dropped. This method non-optionally requires a "healing" step, recovering the downstream performance with a small number of QLoRA finetuning steps \citep{dettmers2024qlora}.

Our hypothesis extends theirs, and posits that for the task of generating text embeddings, the final layer is also not necessary. 
Our pruning experiments are conducted with the percentage pruned $p$, between $0\%$ (all layers intact) and $100\%$ (all layers removed). Given a pruning percentage and a total number of layers $n$, the new number of layers $n^*$ is calculated as
$$
n^* = \lfloor n \times (1 - p) \rfloor
$$

Given a model and its configuration, this straightforward procedure can be integrated with modern LLM implementations with just three lines of code:
\begin{lstlisting}[language=Python]
n = int(config.num_hidden_layers * (1-p))
model.layers = model.layers[:n]
config.num_hidden_layers = n
\end{lstlisting}

We then conduct supervised contrastive training, as with prior work on converting LLMs to text encoders. In lieu of an explicit healing step, we hypothesize that the aforementioned training acts as such. Thus, no additional or separate training is necessary to execute our method. 
 
\section{Experiments}

\begin{figure*}[t]
  \includegraphics[width=\textwidth]{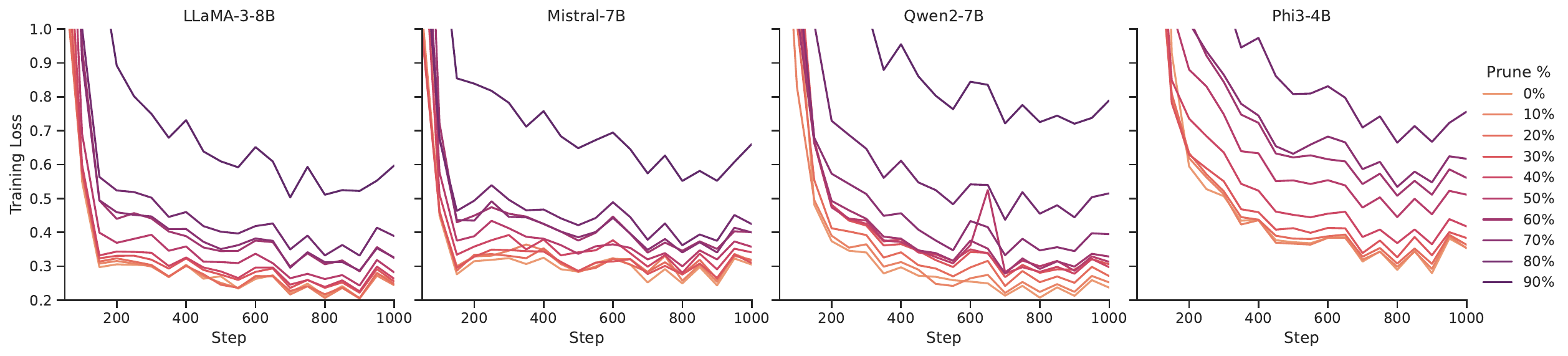}
  \caption{The training loss curves for each model at different pruning percentages.}
  \label{fig:training_curves}
\end{figure*}

\begin{figure}[t]
  \includegraphics[width=\columnwidth]{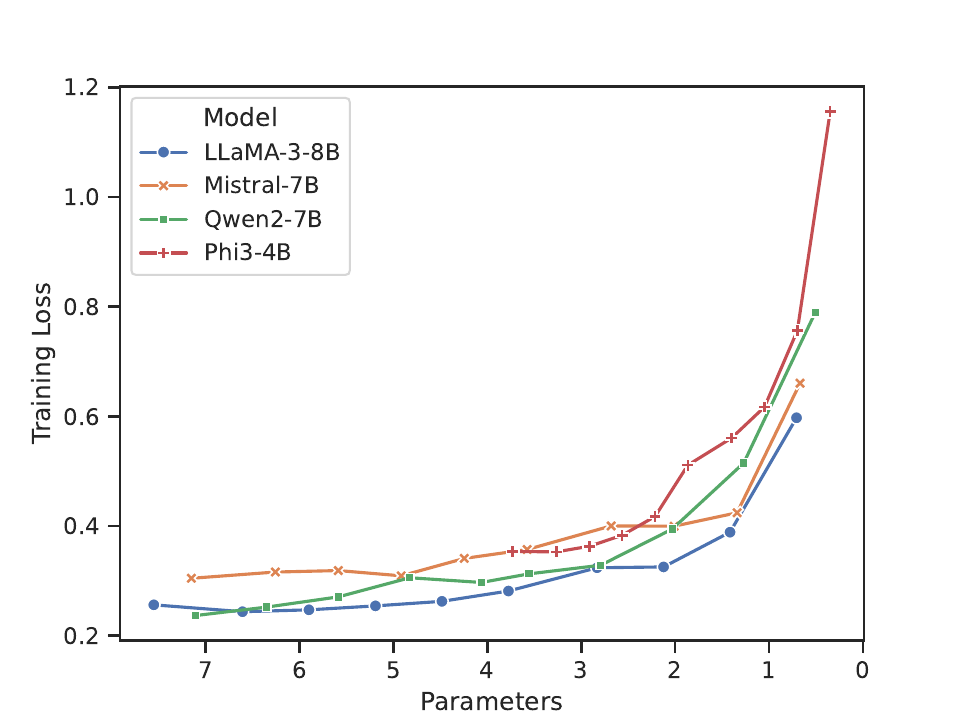}
  \caption{The final loss values at the end of training across different pruning percentages.}
  \label{fig:final_loss_params}
\end{figure}

\subsection{General Setup}

For our experiments, we chose four instruct-tuned decoder-only LLMs across different families ranging from 3.7B to 7.5B: LLaMA-3-8B (\texttt{Meta-Llama-3-8B-Instruct}, \citealp{dubey2024llama3herdmodels}), Mistral-7B (\texttt{Mistral-7B-Instruct-v0.2}, \citealp{jiang2023mistral7b}), Qwen2-7B (\texttt{Qwen2-7B-Instruct}, \citealp{yang2024qwen2technicalreport}), and Phi3-4B (\texttt{Phi-3-mini-4k-instruct}, \citealp{abdin2024phi3technicalreporthighly}). These model families were chosen due to their widespread use in open-source communities and LLM literature.  As we are chiefly concerned with the effects of pruning, we opt for no modification to the LLM architecture itself. We use weighted mean pooling \citep{Muennighoff2022-SGPT} to generate embeddings from the outputs of the LLM as it is straightforward to implement and outperforms other pooling measures when paired with causal attention \citep{Muennighoff2022-SGPT, behnamghader_llm2vec_2024}. 

We also conduct supervised contrastive finetuning, known to outperform unsupervised finetuning and the zero-shot setting, and considered to be an integral part of effectively utilizing LLMs as embedding models \citep{behnamghader_llm2vec_2024, Muennighoff2022-SGPT, jiang-etal-2023-prompteol}. We use the replication of the public portion of the E5 dataset \citep{wang-etal-2024-improving-text}, curated by \citet{springer_repetition_2024}, as the training dataset. Consisting of
approximately 1.5 million samples, it is a multilingual compilation of various retrieval datasets, meant for supervised contrastive training of embedding models. In accordance, we use contrastive loss with hard negatives and in-batch negatives \citep{springer_repetition_2024, behnamghader_llm2vec_2024}. Further details on the dataset and training are provided in Appendix \ref{appendix:training}.

All experiments were conducted on a single A100 GPU, reinforcing the accessible nature of our proposed procedure.

\subsection{Zero-shot Loss Evolution Over Layers}
\label{sec:loss_evo}

\begin{figure*}[t]
  \includegraphics[width=\textwidth]{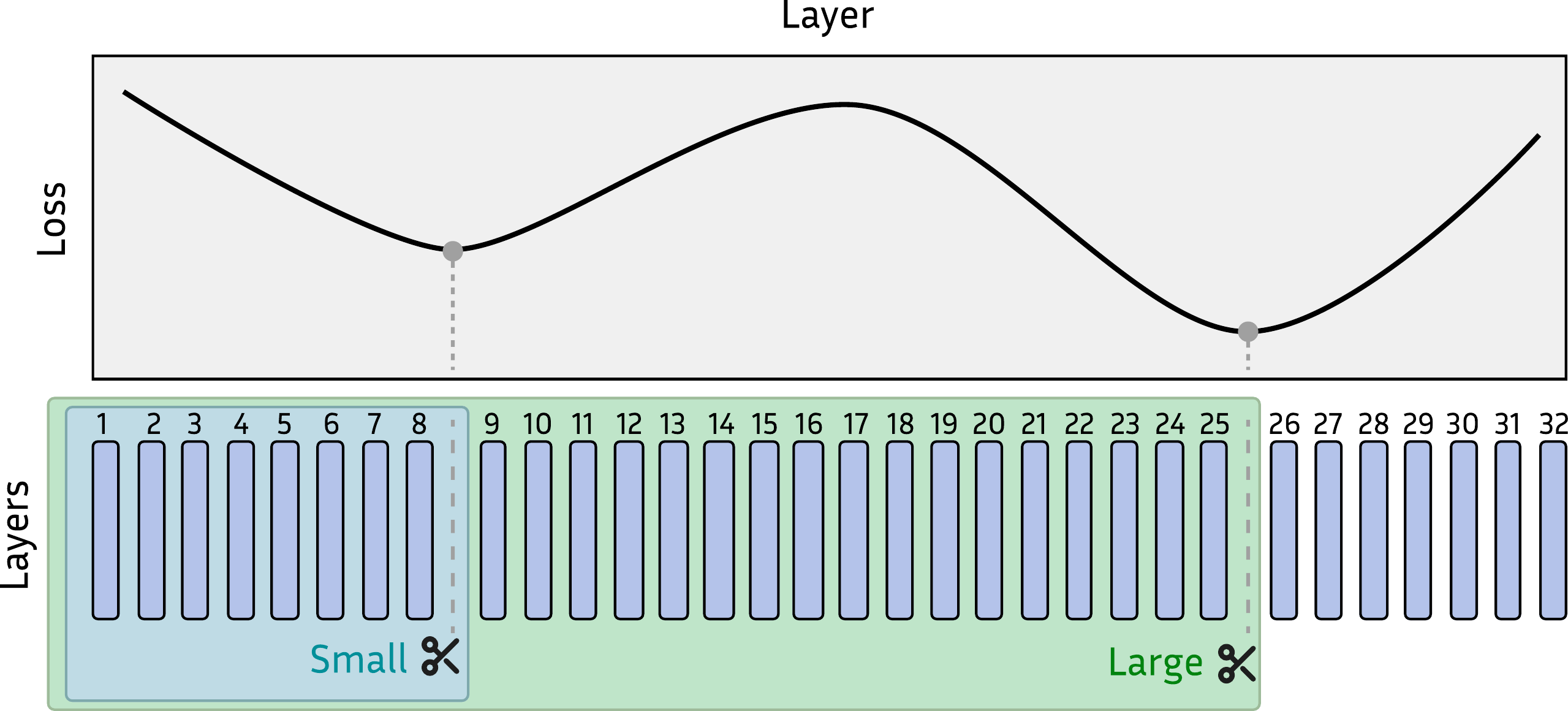}
  \caption{A simplified illustration of L\textsuperscript{3}Prune. The initial loss of the representation of each layer is found, and the two minima before and after 50\% of the model correspond to the layers to prune to in the two configurations, \texttt{small} and \texttt{large}. }
  \label{fig:l3prune}
\end{figure*}

As a preliminary test of our hypothesis—that an LLM can form performant text representations even before reaching the final layer—we first calculate how well the output of each layer of the model performs as an embedding. We note that this is equivalent to a zero-shot setting. As we are interested in a comparative measure between layers intra-model, the loss as a metric is sufficient. We take a random sampling of 1280 tuples from the training dataset and calculate the embeddings via weighted mean pooling of the outputs of each layer. Then, the loss is calculated and averaged per layer. We find that the loss values converge fairly quickly, and so 1280 samples are sufficient for our purposes. The results are aggregated in Figure \ref{fig:loss_evo}.

The loss for all four models follow a similar curve: an initial drop to around layer 5-10, a subsequent rise around layer 15, and then a slower drop up to layer 22-25, where it rises again by the end with layer 28-32. While the specifics of the ways in which LLM representations evolve are not well-understood, these results suggest that the early layers of the model are generally focused on representation, while the final layers transform the representation into the specific probability distribution for the next token. Regardless of the underlying dynamics, the drop-rise-drop curve is consistent across model sizes and families in our experiments.

We expect that training will transform the shape of this layerwise evolution considerably. We also have little reason to expect that the final downstream performance of layer-dropped models will be accurately modeled by the effectiveness of these initial representations. However, we posit that these initial loss curves also reveal optimal starting points for pruning. The minima of these curves indicate layers where the text embeddings are best-optimized, making them good candidates for pruning without significant performance loss. 

Inspired by these findings, we consider the following procedure for pruning: find the two minima in the layer-loss curve before and after 50\% of the layers (the low point of the two drops). We hypothesize that pruning up to these layers provides us with two models: a smaller model with degraded but reasonable performance, and a larger model whose performance is close to the original. This procedure would thus produce two text embedding model variants from an LLM, each variant usable in different circumstances. The two aforementioned models are termed \texttt{large} and \texttt{small} in the following sections. We term this method \textbf{L}LM \textbf{L}ayerwise \textbf{L}oss \textbf{P}runing, or \textbf{L\textsuperscript{3}Prune} for short. Figure \ref{fig:l3prune} shows a simple illustration of the process.

\subsection{Supervised Training}
\label{sec:supervised_training}

In order to verify the general efficacy of our hypothesis—that LLMs can form effective text representations even before reaching their deeper layers—we conduct training on pruned LLMs to convert them into effective text encoders. We keep the training procedure fairly straightforward: supervised contrastive learning for 1000 steps with LoRA modules \citep{hu2022lora}. Other hyperparameters are detailed in Appendix \ref{appendix:training:hyperparams}. We first test a range of pruning percentages from 10\% to 90\% at 10\% intervals. Figure \ref{fig:training_curves} shows the training loss for all models and pruning percentages. We note that the training loss curves all generally follow the same shape, indicating stability in training even with the modified architecture. 

Figure \ref{fig:final_loss_params} shows the final loss in relation to the pruned model parameters, with each marked point representing a model pruned by an additional 10\%. The final loss values for each model follow a straightforward trajectory with increasing pruning percentage: minimal increases up to 30-40\%, with larger increases as the pruning percentage hits 90\%. Notably, we find that the final loss of different models correlates more with the final parameter count after pruning than with the percentage of layers retained. This suggests that the parameter count is a more significant factor in determining the effectiveness of a pruned model than simply the proportion of layers kept. 

If we presume that training loss correlates well with downstream accuracy for text embedding, we can make a series of predictions from an analysis of the plots:
\begin{itemize}
    \item Performance always degrades sharply as the parameter count approaches and goes below 1 billion.
    \item In contrast, performance degrades little even with 30-50\% pruning. LLaMA-3-8B degrades minimally up to 40-50\%, Mistral-7B up to 30-40\%, and Phi3-4B up to 20-30\%. Qwen2-7B degrades more at low pruning percentages, but remains stable between 30-60\%.
    \item Even at high pruning percentages, model performance degrades at a reasonable rate. Models can likely be pruned up to 2 billion parameters while still producing viable embeddings.
\end{itemize}

\subsection{Simple Pruning Evaluation}
\label{sec:evaluation}

\begin{figure}[t]
  \includegraphics[width=\columnwidth]{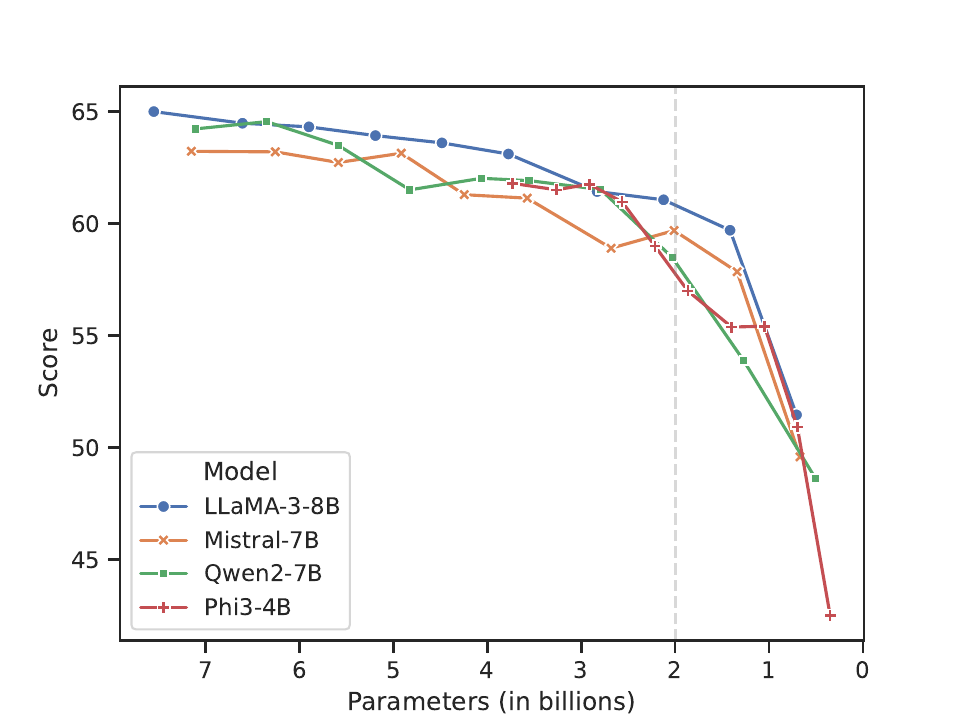}
  \caption{The MTEB (15 task subset) scores with respect to the number of model parameters.}
  \label{fig:mteb_parameters}
\end{figure}

To validate the predictions made from the training loss, we evaluate the models at various pruning percentages on downstream text embedding tasks. Specifically, to speed up evaluation, we opt for the 15-task subset of the Massive Text Embedding Benchmark (MTEB, \citealp{muennighoff-etal-2023-mteb}) collected and used by \citet{behnamghader_llm2vec_2024}. The subset, which we term MTEB-15 for clarity, covers representative tasks from the full 56 tasks in MTEB, including tasks from each category with almost the same proportion to prevent bias. Further details are provided in Appendix \ref{appendix:mteb15}.

In accordance with previous work \citep{behnamghader_llm2vec_2024, springer_repetition_2024,wang-etal-2024-improving-text}, we evaluate with task-specific instructions. We use the same instructions as \citet{wang-etal-2024-improving-text}. The instructions can be found in Appendix Table \ref{tab:mteb_instructions}. Following \citet{behnamghader_llm2vec_2024}, for symmetric tasks, the same instruction is used for the query and the document. Instruction tokens are excluded from the final pooling.  

Figure \ref{fig:mteb_parameters} shows the impact of pruning on MTEB-15 results across a range of pruning percentages. We plot with respect to the number of parameters as opposed to relative pruning percentages because parameter count correlates better with the score. We can see that the training loss and MTEB-15 score also roughly correlate. This confirms that our predictions in Section \ref{sec:supervised_training}, based on the supervised training loss, are fairly accurate. 

LLama at 50\% pruning (3.77B) is only degraded by $-1.89$, still providing a strong performance of $63.10$. Even at 80\% pruning (1.41B), it performs at a reasonable $59.69$. Mistral's performance decrease is an almost negligible $-0.08$ up to 30\% (4.91B). Qwen's performance \textit{increases} by +0.32 with a pruning of 10\%. It drops distinctly at 30\% pruning. However, it stabilizes at a reasonable $61.51$ up till 60\% (2.79B). Phi degrades negligibly up to 20\% pruning (2.91B) with $-0.03$, and $-0.53$ at 30\% (2.56B). Higher pruning percentages degrade it significantly, as the model parameter count decreases below the 2 billion mark. 

Our results correspond roughly with those of \citet{gromov_unreasonable_2024}: sharp transitions in performance around 45\%-55\% for models in the Llama family, 35\% for Mistral, 25\% for Phi, and 20\% for Qwen. However, instead of a sharp transition to near-random performance, we observe a steady but reasonable decline even at higher pruning percentages. In general, we only observe a significant decline in performance as model size goes below roughly 2 billion parameters. These results also correlate roughly with previous findings by \citet{jiang-etal-2023-prompteol}, who investigated LLM-based sentence embedding models between 125M to 66B parameters and found diminishing returns at parameter counts over 2B.

We can derive some general insights from these experiments. For one, the resilience of a model to pruning is not entirely consistent across families and sizes. Thus, model-specific experimentation may be required. However, in general, models can be pruned 10-30\% with minimal drop in downstream performance. Further, higher pruning percentages up to 80\% still yield reasonably effective embedding models. 

We note that LLaMa-3-8B at 50\% pruning, with 3.77B parameters, outperforms an unpruned Phi3-4B at 3.73B parameters. In conjunction with our other results, we suggest that given a compute/memory budget, simply dropping layers of a high-performing LLM may be a superior and significantly simpler strategy than training a smaller LM that fits the budget.

\begin{table*}[t]
  \centering
  \small
  \begin{tabular}{lcccccccc}
  \toprule
  & \multicolumn{2}{c}{\textbf{LLaMA-3-8B}} & \multicolumn{2}{c}{\textbf{Mistral-7B}} & \multicolumn{2}{c}{\textbf{Qwen2-7B}} & \multicolumn{2}{c}{\textbf{Phi3-4B}} \\
  \cmidrule(lr){2-3} \cmidrule(lr){4-5} \cmidrule(lr){6-7} \cmidrule(lr){8-9}
  & \textbf{Large} & \textbf{Small} & \textbf{Large} & \textbf{Small} & \textbf{Large} & \textbf{Small} & \textbf{Large} & \textbf{Small} \\
  \midrule
  \textbf{Layers} & 25 \textcolor{teal}{(-7)} & 5 \textcolor{teal}{(-27)} & 22 \textcolor{teal}{(-10)} & 8 \textcolor{teal}{(-24)} & 25 \textcolor{teal}{(-3)} & 10 \textcolor{teal}{(-18)} & 25 \textcolor{teal}{(-7)} & 8 \textcolor{teal}{(-24)} \\
  \textbf{Params} & 5.9 \textcolor{teal}{(78\%)} & 1.18 \textcolor{teal}{(16\%)} & 4.92 \textcolor{teal}{(69\%)} & 1.79 \textcolor{teal}{(25\%)} & 6.35 \textcolor{teal}{(89\%)} & 2.54 \textcolor{teal}{(36\%)} & 2.91 \textcolor{teal}{(78\%)} & 0.93 \textcolor{teal}{(25\%)} \\
  \textbf{Score} & 63.5 \textcolor{red}{(-1.5)} & 58.1 \textcolor{red}{(-6.9)} & 63.1 \textcolor{red}{(-0.1)} & 59.0 \textcolor{red}{(-4.2)} & 64.5 \textcolor{teal}{(+0.3)} & 60.9 \textcolor{red}{(-3.3)} & 61.7 \textcolor{red}{(-0.1)} & 55.5 \textcolor{red}{(-6.3)} \\
  \bottomrule
  \end{tabular}
  \caption{Comparison of \texttt{large} and \texttt{small} variants across various models, including number of layers, parameters, and MTEB scores. Changes from the full model are provided in paranthesis.}
  \label{tab:family_results}
\end{table*}

\subsection{L\textsuperscript{3}Prune Evaluation}
\label{sec:l3prune}

As mentioned in Section \ref{sec:loss_evo}, we hypothesize that the minima in the layer-loss curve before and after the midpoint are particularly effective points for pruning. We prune to those particular layers and conduct the same training and evaluation as described in Sections \ref{sec:supervised_training} and \ref{sec:evaluation}. Table \ref{tab:family_results} aggregates the results across base models for the two resulting prune configurations, termed \texttt{small} and \texttt{large}. It also shows the particular layer numbers and parameter counts.

The results are consistent with our previous findings. The \texttt{small} models generally perform worse than the full-sized models, with performance drops ranging between $-4.4$ and $-6.9$. However, at 16\%-36\% of their original size (84\%-64\% pruning), the models are proportionally compute and memory-efficient in exchange for the dropped performance. The \texttt{large} models, on the other hand, perform almost as well as the unpruned models, with only a slight drop in performance, while pruned to 69\%-89\% (31\%-11\% pruning). As we have seen before, Qwen2-7B's performance \textit{increases} slightly with pruning, and both Mistral-7B and Phi3-4B's performance drops are negligible. LLaMA-3-8B's performance drops by $-1.4$ points but still remains a fairly strong $63.5$.

Combined with the results from Section \ref{sec:evaluation}, we can see that the layers picked by L\textsuperscript{3}Prune are generally optimal. For instance, Mistral-7B, Qwen2-7B, and Phi3-4B show strong performances up to 30\%, 10\%, and 20\% pruning respectively, and the layers corresponding to those pruning percentages are exactly the layers pinpointed by L\textsuperscript{3}Prune for the \texttt{large} variant. As LLaMa-3-8B's performance decrease remains fairly consistent when pruning below 50\%, we infer that there is no particularly optimal point for pruning. Similarly, the \texttt{small} variants are pruned up to the point before each model's performance drops drastically—roughly 85\% for LLAMA-3-8B, 75\% for Mistral-7B, 65\% for Qwen2-7B, and 75\% for Phi3-4B. 

Based on these results, we can conclude that the layerwise loss evolution of a model can be used to effectively pick optimal points for pruning. The resulting variants can be used to provide a range of models with different performance and efficiency trade-offs. The \texttt{large} models are particularly effective, with a negligible drop (or even an increase) in performance for a significant reduction in size. The \texttt{small} models can be used for resource-constrained settings, with reasonable performance.

We further note that the training of the \texttt{small} variants required only 23.6 GB of VRAM at maximum, and the layerwise loss curves can be calculated with less than 17 GB of VRAM. The training is only conducted for 1000 steps and takes less than an hour on average using an A100 GPU. Thus, \texttt{small} variant models can be trained on consumer-grade GPUs, making it accessible to open-source and practitioner communities. Further details on training times are given in Appendix \ref{appendix:training:times}.

\section{Conclusion}
In this work, we presented a simple and effective pruning approach to convert LLMs into lightweight, performant text embedding models. By dropping the last $p\%$ layers of the model, we achieved significant reductions in model size and inference time, with minimal impact on text embedding tasks. Our procedure is straightforward to implement in pipelines converting LLMs to text encoders, and requires no additional training, providing smaller models at no cost. Based on the initial model loss, we also proposed L\textsuperscript{3}Prune, a method to pinpoint optimal layers to prune to, providing an efficient strategy for pruning without extensive experimentation. We demonstrated that significant pruning—up to 31\%—can be conducted with a negligible performance loss, and substantial pruning—up to 84\%—can still produce effective models. Overall, our results show that LLMs are significantly overparameterized for text embedding tasks, and can be pruned with minimal performance loss. 

\section{Limitations}
Our work only takes into consideration the supervised finetuning setting for utilizing LLMs as text encoders, as this is the most common and generally effective. Further, our results may not hold with extensive modifications to the architecture or training process. Lastly, even with extensive pruning, our smallest models are still generally larger than traditionally trained encoder-only models.
However, the inexpensive finetuning required and the performance advantages offered by LLM-based text encoders motivates pruning as a viable alternative in resource-constrained settings. 

\section{Ethical Considerations}
Our work provides an effective and efficient method to produce optimized text embedding models from LLMs. As we mentioned in Section \ref{sec:l3prune}, our method is memory and compute-efficient, and can be conducted on consumer-grade GPUs, making it accessible for a wider audience of practitioners and academics. However, this also enhances potential issues with misuse, lowering the bar for malicious actors to train and host embedding models. Regardless, embedding models in general have significantly fewer avenues for malicious behaviour in comparison to e.g. generative LLMs.

\bibliography{anthology,custom}

\appendix

\section{Training}
\label{appendix:training}
\subsection{Dataset}
\label{appendix:training:dataset}

The dataset we use consists of ELI5 (sample ratio 0.1, \citealp{fan-etal-2019-eli5}), HotpotQA \citep{yang-etal-2018-hotpotqa}, FEVER \citep{thorne-etal-2018-fever}, MIRACL \citep{zhang-et-all-2023-MIRACL}, MS-MARCO passage ranking (sample ratio 0.5) and document ranking (sample ratio 0.2, \citealp{bajaj-etal-2018-MSMARCO}), NQ \citep{karpukhin-etal-2020-DPR}, SNLI \citep{bowman-etal-2015-large}, MNLI \citep{williams-etal-2018-broad}, SQuAD \citep{rajpurkar-etal-2016-squad},
TriviaQA \citep{joshi-etal-2017-triviaqa},
Quora Duplicate Questions\footnote{\url{https://quoradata.quora.com/First-Quora-Dataset-Release-Question-Pairs}} (sample ratio 0.1),
Mr- TyDi \citep{zhang-etal-2021-mr}, DuReader \citep{he-etal-2018-dureader}, and T2Ranking (sample ratio 0.5, \citealp{t2ranking}).
The instructions used for each dataset can be found in Table \ref{tab:finetuning_instructions}.

\subsection{Hyperparameters}
\label{appendix:training:hyperparams}

All models are trained with LoRA rank $r = 16$ and use brain floating point (\texttt{bfloat16}) precision, gradient checkpointing, and FlashAttention-2 \cite{dao2023flashattention2} to optimize GPU memory consumption. Training is conducted with a batch size of 64 for 1000 steps, with gradient accumulation over 1 step, and a maximum sequence length of 512 tokens. The Adam optimizer is employed with a learning rate of $2 \times 10^{-4}$ and a linear warm-up over the first 300 steps.

\subsection{Training Time}
\label{appendix:training:times}

\begin{figure}[h]
  \includegraphics[width=\columnwidth]{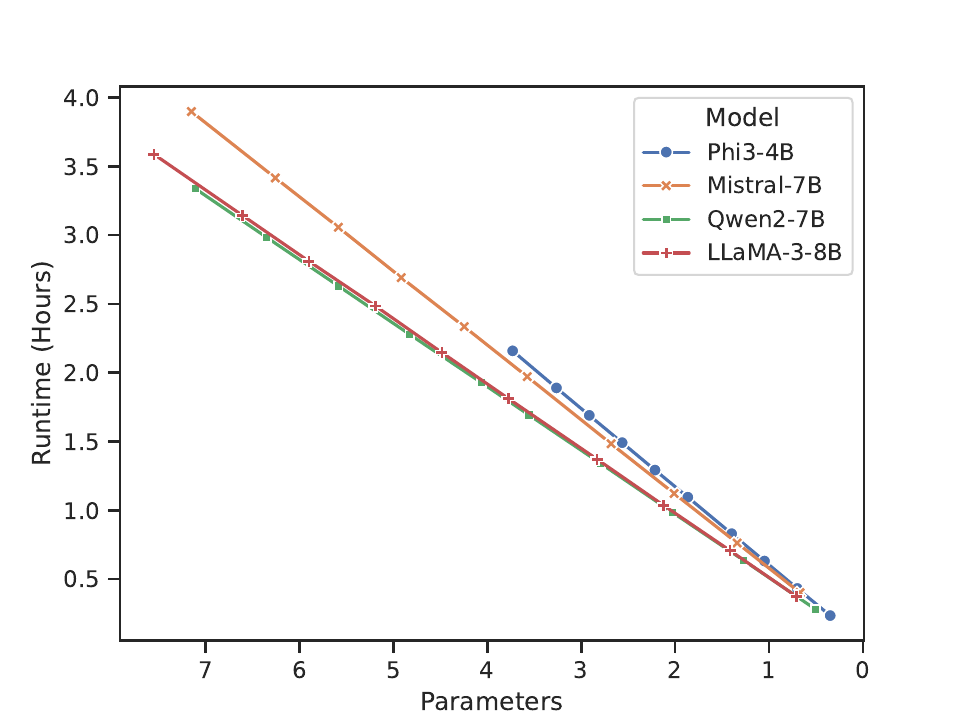}
  \caption{The total training time taken for all models at different pruning percentages.}
  \label{fig:all_training_times}
\end{figure}

\begin{table}[h!]
\centering
\begin{tabular}{p{3cm}p{1.5cm}p{1.5cm}}
\toprule
 & \textbf{Large} & \textbf{Small} \\ \midrule
\textbf{LLaMA-3-8B} & 2h 48m & 35m \\ 
\textbf{Mistral-7B} & 2h 41m & 56m \\ 
\textbf{Qwen2-7B} & 3h 1m & 1h 14m \\ 
\textbf{Phi3-4B} & 1h 40m & 33m \\ \bottomrule
\end{tabular}
\caption{Training time for the variants produced by L\textsuperscript{3}Prune.}
\label{tab:traintime}
\end{table}

Table \ref{tab:traintime} shows the time taken to train the two variants (\texttt{large} and \texttt{small}) provided by L\textsuperscript{3}Prune for each model. Figure \ref{fig:all_training_times} shows the training time for the models pruned at different pruning percentages, with respect to total parameter count. As we expect, the time taken to train a pruned model is linear to the pruning percentage, and corresponds roughly to the total parameter count. All models were trained on a single NVIDIA A100 GPU. Including evaluation, we estimate that all experiments took a total of 200 GPU hours.

\section{Massive Text Embeddings Benchmark (MTEB)}
\label{appendix:mteb}
\subsection{MTEB subset details}
\label{appendix:mteb15}

MTEB encompasses a diverse array of embedding tasks varying in size, making a full evaluation quite time-consuming—it takes over 160 hours for a full-sized 7B model, such as Qwen2-7B, on an A100 GPU. To expedite our analysis, we use a representative subset of 15 tasks from MTEB, selected and used by \citet{behnamghader_llm2vec_2024}, detailed in Table \ref{tab:mteb-subset}. This subset includes tasks from each category in proportions closely matching those of the full MTEB.

\begin{table}[ht]
    \centering
    \small
    \begin{tabular}{ll}
    \toprule
    \textbf{Category} & \textbf{Dataset} \\
    \midrule
    \multirow{3}{*}{Retrieval (3)} & SciFact \\
    & ArguAna \\
    & NFCorpus \\ \midrule
    \multirow{2}{*}{Reranking (2)} & StackOverflowDupQuestions \\
    & SciDocsRR\\ \midrule
    \multirow{3}{*}{Clustering (3)} & BiorxivClusteringS2S \\
    & MedrxivClusteringS2S \\
    & TwentyNewsgroupsClustering \\ \midrule
    Pair Classification (1) & SprintDuplicateQuestions \\ \midrule
    \multirow{3}{*}{Classification (3)} & Banking77Classification \\
    & EmotionClassification \\
    & MassiveIntentClassification \\ \midrule
    \multirow{3}{*}{STS (3)} & STS17 \\
    & SICK-R \\
    & STSBenchmark \\ \midrule
    SummEval (0) & -\\ \midrule
    Overall & 15 datasets \\
    \bottomrule
    \end{tabular}
    \caption{MTEB-15, the subset of MTEB tasks used for our work.}\label{tab:mteb-subset}
\end{table}


\subsection{MTEB instructions}
\label{appendix:mteb_instructions}

For evaluation on MTEB-15, we use the instructions from \citet{wang-etal-2024-improving-text}, also used by \citet{behnamghader_llm2vec_2024}. The list of instructions for each task is listed in Table \ref{tab:mteb_instructions}. 

\section{Licenses}

All four models we used are available for research purposes—LLaMA-3-8B is under its own permissive license, Mistral-7B and Qwen2-7B are under Apache License 2.0, and Phi3-4B is under MIT License. MTEB and the tasks it includes are provided under the Apache License 2.0. We overview the licenses of all datasets used in training below:
\begin{itemize}
    \item ELI5: Provided under no specified license, available for research purposes.
    \item HotpotQA: Provided under CC BY-SA 4.0.
    \item FEVER: Provided under CC BY-SA 3.0.
    \item MIRACL: Provided under Apache License 2.0.
    \item MS-MARCO: Provided under no specific license, available for non-commercial research purposes.
    \item Natural Questions (NQ): Provided under CC BY 4.0.
    \item Stanford Natural Language Inference (SNLI): Provided under CC BY-SA 4.0.
    \item Multi Natural Language Inference (MNLI): Provided under a combination of permissive licenses, elaborated by \citet{williams-etal-2018-broad}.
    \item SQuAD: Provided under CC BY-NC 4.0.
    \item TriviaQA: Provided under Apache License 2.0.
    \item Quora Duplicate Questions: Provided under no specified license, available for non-commercial purposes.
    \item Mr. TyDi: Provided under Apache License 2.0
    \item DuReader: Provided under Apache License 2.0
    \item T2Ranking: Provided under Apache License 2.0
\end{itemize}

\begin{table*}[ht]
\begin{tabular}{p{0.3\linewidth}p{0.65\linewidth}}
\toprule
\textbf{Task Name} & \textbf{Instruction} \\
\midrule
Banking77Classification & Given a online banking query, find the corresponding intents  \\
EmotionClassification &  Classify the emotion expressed in the given Twitter message into one of the six emotions: anger, fear, joy, love, sadness, and surprise  \\
MassiveIntentClassification & Given a user utterance as query, find the user intents  \\
BiorxivClusteringS2S & Identify the main category of Biorxiv papers based on the titles  \\
MedrxivClusteringS2S & Identify the main category of Medrxiv papers based on the titles  \\
TwentyNewsgroupsClustering & Identify the topic or theme of the given news articles  \\
SprintDuplicateQuestions & Retrieve duplicate questions from Sprint forum  \\
SciDocsRR & Given a title of a scientific paper, retrieve the titles of other relevant papers  \\
StackOverflowDupQuestions & Retrieve duplicate questions from StackOverflow forum  \\
ArguAna & Given a claim, find documents that refute the claim  \\
NFCorpus & Given a question, retrieve relevant documents that best answer the question  \\
SciFact & Given a scientific claim, retrieve documents that support or refute the claim  \\
STS* & Retrieve semantically similar text.  \\
\bottomrule
\end{tabular}
\caption{Instructions used for evaluation on the MTEB benchmark.
``STS*'' refers to all the STS tasks.} \label{tab:mteb_instructions}
\end{table*}

\begin{table*}[ht]
\centering
\begin{tabular}{p{0.15\linewidth}p{0.8\linewidth}}
\toprule
\textbf{Dataset} & \textbf{Instruction(s)} \\
\midrule
SNLI \& MNLI & Given a premise, retrieve a hypothesis that is entailed by the premise \\
    & Retrieve semantically similar text \\
DuReader & Given a Chinese search query, retrieve web passages that answer the question \\
ELI5 & Provided a user question, retrieve the highest voted answers on Reddit ELI5 forum \\
FEVER & Given a claim, retrieve documents that support or refute the claim \\
HotpotQA & Given a multi-hop question, retrieve documents that can help answer the question \\
MIRACL & Given a question, retrieve Wikipedia passages that answer the question \\
MrTyDi & Given a question, retrieve Wikipedia passages that answer the question \\
MSMARCO Passage & Given a web search query, retrieve relevant passages that answer the query \\
MSMARCO Document & Given a web search query, retrieve relevant documents that answer the query \\
NQ & Given a question, retrieve Wikipedia passages that answer the question \\
QuoraDuplicates & Given a question, retrieve questions that are semantically equivalent to the given question \\
 & Find questions that have the same meaning as the input question \\
SQuAD & Retrieve Wikipedia passages that answer the question \\
T2Ranking & Given a Chinese search query, retrieve web passages that answer the question \\
TriviaQA & Retrieve Wikipedia passages that answer the question \\
\bottomrule
\end{tabular}
\caption{Instructions used for each of the E5 datasets.}
\label{tab:finetuning_instructions}
\end{table*}

\end{document}